\documentclass[journal]{IEEEtran}

\usepackage{cite}
\usepackage{amsmath,amssymb,amsfonts}
\usepackage{algorithmic}
\usepackage{graphicx}
\usepackage{textcomp}
\usepackage{xcolor}
\usepackage[utf8]{inputenc} 
\usepackage[T1]{fontenc}    
\usepackage{hyperref}       
\usepackage{url}            
\usepackage{booktabs}       
\usepackage{amsfonts}       
\usepackage{nicefrac}       
\usepackage{microtype}      
\usepackage{comment}
\usepackage{color}
\usepackage{mathtools}
\usepackage{caption}
\usepackage{subcaption}
\usepackage{multirow}
\usepackage{booktabs}
\usepackage{diagbox}
\usepackage{bm}
\usepackage{float}
\usepackage{stfloats}

\begin{document}
\title{Convolutional Neural Network optimization via Channel Reassessment Attention module}

\author{Yutao~Shen and Ying Wen%
	\IEEEcompsocitemizethanks{
	\IEEEcompsocthanksitem {Y. Shan is with the School of Computer Science and Technology, East China Normal University, Shanghai 200062, China \protect\\
			E-mail: 51184506069@stu.ecnu.edu.cn.}
	\IEEEcompsocthanksitem {Y. Wen is with the School of Communications and Electronic Engineering, East China Normal University, Shanghai
			200241, China \protect\\
			E-mail: ywen@cs.ecnu.edu.cn.  \protect\\
			Corresponding author: Ying Wen}} \protect\\
}


\maketitle

\begin{abstract}
	The performance of convolutional neural networks (CNNs) can be improved by adjusting the interrelationship between channels with attention mechanism. 
	However, attention mechanism in recent advance has not fully utilized spatial information of feature maps, 
	which makes a great difference to the results of  generated channel attentions. 
	In this paper, we propose a novel network optimization module called \textit{Channel Reassessment Attention} (CRA) module which uses channel attentions with spatial information of feature maps to enhance representational power of networks.
	We employ CRA module to assess channel attentions based on feature maps in different channels, then the final features are refined adaptively by product between channel attentions and feature maps.
	CRA module is a computational lightweight module and it can be embedded into any architectures of CNNs.
	The experiments on ImageNet, CIFAR and MS COCO datasets demonstrate that the embedding of CRA module on various networks effectively improves the performance under different evaluation standards.
\end{abstract}

\begin{IEEEkeywords}
	Convolutional neural networks, Network optimization, Attention mechanism
\end{IEEEkeywords}

\IEEEpeerreviewmaketitle

\section{Introduction}
\label{sec:introduction}
\IEEEPARstart{C}{onvolutional} neural networks (CNNs) are widely used in visual tasks due to their superior performance in extracting features \cite{Krizhevsky2012,Zeiler2014,Sermanet2014,Ren2017,Shelhamer2017}.
The first proposed AlexNet \cite{Krizhevsky2012} leads to the vigorous development of CNN.
Since then, many different CNN architectures such as GoogLeNet \cite{Szegedy}, VGG \cite{Simonyan2014} and ResNet \cite{He2016} have been proposed. 
GoogLeNet proposes Inception module to increase the width of networks and integrates multiple receptive fields information.
VGG uses more parameters in a simple network architecture. 
ResNet proposes a skip connection to deepen the depth of CNNs.
DenseNet uses dense connections to concatenate feature maps between different layers.
Besides, NAS \cite{Zoph2017,Baker2017} is proposed to search for network architectures with better performance.

In general, for input feature maps in each convolutional layer, CNN model learns the optimal kernel filter parameters to obtain higher hierarchical feature maps. 
Essentially, convolution operation fuses spatial and channel-wise information of input feature maps. 
By stacking a series of convolutional layers, activation layers, batch normalization layers \cite{Ioffe2015} and pooling layers, 
networks can hierarchically capture semantic features from previous feature maps, which represent the original input image.
However, not all feature maps have the same contribution in the propagation of network.
Recent works have considered relationships between channels through attention mechanism to increase the accuracy of CNNs. 
STN \cite{Jaderberg2015} uses spatial information of feature maps to make networks invariant to translation, rotation and other spatial transformations.
SENet \cite{Hu2018} shows that emphasising informative features and suppressing less useful ones can improve the representational power of a network.
SCA-CNN \cite{Chen2017}, BAM \cite{Park2019} and CBAM \cite{Woo2018} combine spatial and channel attentions to refine convolutional features.
In the non-local \cite{Wang2018} block architecture, global self-attention mechanism is used to capture long-range interactions, which improves the performance of CNNs in video classification and object detection tasks.
DANet \cite{Fu2019} and CCNet \cite{Huang2019} capture global feature dependencies in spatial and channel dimensions to achieve better performance in image segmentation task.
AutoPruner \cite{Luo2018} proves that even removing unimportant channels directly does not have much impact on the performance of CNNs. 

When dealing with global information of feature maps, the methods mentioned above either use a global pooling operation , which directly compress a 2D feature map into a scalar without considering the spatial information, or deploy operations with high computational cost to process spatial information of feature maps.
However, spatial information of different feature maps carries visual and semantic differences,
which is crucial to the effectiveness of extracted channel attentions.
Moreover, high computational cost will make networks difficult to train, and limit the application value of networks.
Therefore, how to make full use of spatial information to optimize networks while keep less computational cost is an important issue.

In this paper, we propose a novel network optimization module called \textit{Channel Reassessment Attention} (CRA) module, which utilizes a pooling operation to compress feature maps and then uses global depthwise convolution (GDConv) to assess channel attentions of each compressed feature maps. 
There are two advantages of the proposed CRA module: 
(1) The proposed CRA module captures global-range dependencies 
between spatial positions; 
(2) CRA module is a computationally lightweight module and can be easily embedded into different architectures of CNNs.
In the following section, we evaluate CRA module with various different architectures of CNNs on several common datasets, the experimental results validate that our method is effective.

\begin{figure}[t]
\centering
\includegraphics[width=1.0\linewidth]{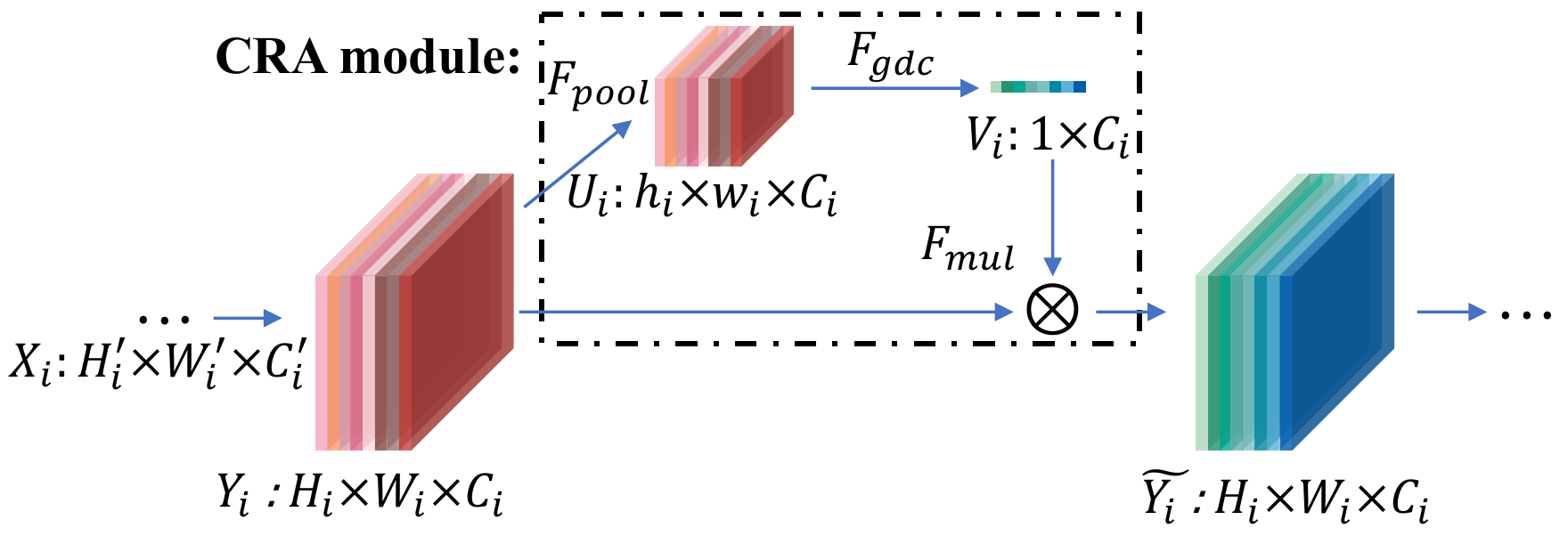}
\caption{The architecture of the proposed CRA module.}
\label{fig:arth}
\end{figure}

\section{Method}
\label{sec:method}
In this section, we will formulate the proposed CRA module, 
which uses channel attentions with spatial information of feature maps to optimize CNNs.

In a CNN, the operation of the $i$-th convolution layer can be defined as the following function: 
\begin{equation}
\begin{aligned}
Y_i=W_i \odot X_i
\label{eq:conv1}
\end{aligned}
\end{equation}
where $X_i=[ x_i^{1},x_i^{2},\dots,x_i^{C_{i}^{'}} ]$ are input feature maps and 
$Y_i=[ y_i^{1},y_i^{2},\dots,y_i^{C_{i}} ]$ are output feature maps, 
while $x_i^{j}$ represents the $j$-th 2D input feature map with the size $\langle H_i^{'},W_i^{'} \rangle$ and 
$y_i^{j}$ represents the $j$-th 2D output feature map with the size $\langle H_{i},W_{i}\rangle$, $C_{i}^{'}$ and $C_{i}$ denote the number of channels of $X_i$ and $Y_i$ respectively, 
$\odot$ refers to convolution symbol and $W_i=[W_i^{1}, W_i^{2},\dots,W_i^{C_{i}} ]$ denote the convolution kernels where $W_i^{j}=[w_{i}^{j,1}, w_{i}^{j,2},\dots,w_{i}^{j,C_{i}^{'}}]$ represents the $j$-th spatial filter kernel. 
For simplicity, we omitt the bias term. 
The equivalent form of Eq.\ref{eq:conv1} can be written as follows: 
\begin{equation}
\begin{aligned}
y_i^{j}=W_i^{j} \odot X_i = \sum_{s=1}^{C_i^{'}} w_i^{j,s} \odot x_i^{s}.
\label{eq:conv2}
\end{aligned}
\end{equation}

In Eq.\ref{eq:conv2}, we can see that feature maps of different channels are generated by different filter kernels.
Since different kernels capture different types of features, feature maps of different channels have different contributions to the performance of networks.
Thus, in the paper, we aim to use channel attentions to strengthen the responses of channels with rich feature information and weaken useless ones so that the networks concentrate more on important features.
A channel attention process can be written as:
\begin{equation}
\begin{aligned}
\widetilde{y_i^{j}}&=y_i^{j} \otimes M(y_i^{j})
\label{eq:att}
\end{aligned}
\end{equation}
where $M$ denotes an operation to extract channel attentions and $\widetilde{y_i^{j}}$ refers to the final refined output.
The richer the information one feature map carries, the greater the value of channel attentions extracted.
To make full use of informative features to improve the performance of networks, we propose a CRA module that includes two parts: compressing spatial information and extracting channel attentions.
Fig.\ref{fig:arth} illustrates the architecture of CRA module, which will be described in detail in the following section.

\subsection{Compression of spatial information}
\label{subsec:compression}
In order to access channel attentions, we need to evaluate every feature map $y_i^{j}$. 
However, it is difficult to perform this operation directly due to oversized spatial dimension and complex feature information. 
This issue is particularly severe in the previous layers because the spatial size of feature maps in the previous layers is generally larger than that in the later layers, which directly leads to excessive computing cost.
To solve this problem, we use the average pooling to compress the spatial size of  $y_i^{j}$ into a smaller size while keep sufficient spatial information. 

As show in Fig.\ref{fig:arth}, let $F_{pool}$ represent the average pooling and $U_i$ denote the outputs of $Y_i$ through the function as:
\begin{equation}
\begin{aligned}
U_i=[u_i^{1},&u_i^{2},\dots,u_i^{C_{i}}]=F_{pool}(Y_i)
\\&u_i^{j}=F_{pool}(y_i^j)
\end{aligned}
\end{equation}
where $u_i^{j}$ has the size of $\langle h_{i},w_{i}\rangle$ and $h_i \le H_i , w_i \le W_i$. 
By employing a reasonable value of $\langle h_{i},w_{i}\rangle$, we can retain most of the information of the original $y_i^{j}$ and minimize the subsequent computational cost. 
In the experimental section we will discuss the influence of different $\langle h_{i},w_{i}\rangle$ on the performance of networks. 

\subsection{Extraction of channel attentions}
\label{subsec:extract}
\begin{figure}[t]
	\centering
	\includegraphics[width=1.0\linewidth]{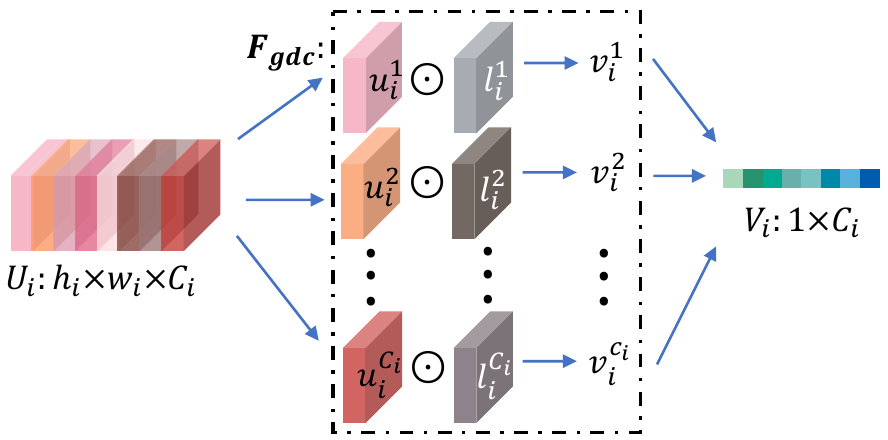}
	\caption{The diagram of GDConv.}
	\label{fig:gdc}
\end{figure}

After we obtain $U_i$ from above operation, we extract channel attention from each $u_i^{j}$.
However, we have $h_i \times w_i$ values in $u_i^{j}$, and these spatial information have different impacts on the attention of $u_i^{j}$. 
It is difficult to extract channel attentions artificially from so many complex feature maps. 
And the extraction operation also subjects to network complexity.
To tackle the two issues, we introduce a global depthwise convolution (GDConv). 
There are three advantages using GDConv: 
(1) Channel attentions can be directly extracted from all spatial information in feature maps through global convolution; 
(2) The computational cost is reduced significantly by depthwise convolution; 
and (3) Channel attentions corresponding to different channels are extracted independently without affecting each other.
Therefore, using GDConv, networks can learn the optimal kernel parameters by iteration to respond to all spatial information in feature maps while keep computational cost as low as possible. 
Fig.\ref{fig:gdc} shows the diagram of GDConv. 

In Fig.\ref{fig:gdc}, $F_{gdc}$ represents GDConv, $L_i=[l_i^{1}, l_i^{2}, \dots, l_{i}^{C_{i}}]$ denotes kernels corresponding to $F_{gdc}$, where the size of $ l_i^{j}$ is $\langle h_{i},w_{i}\rangle$, which equals to the spatial size of $u_i^{j}$.
Let $V_i$ denotes channel attentions of $U_i$,
then $V_i$ can be defined as:
\begin{equation}
\begin{aligned}
V_i=[v_i^{1},v_i^{2}&,...v_i^{C_{i}}]=F_{gdc}(U_i)
\\v_i^{j}&=\sigma (l_i^{j} \odot u_i^{j})
\end{aligned}
\end{equation}
where $\sigma$ represents the sigmoid function.
Note that $l_i^{j}$ only acts on $u_i^{j}$ and they have the same size, thus the result after GDConv operation is a scalar, then by the sigmod function, the result $v_i^{j}$ is taken as channel attention of $u_i^{j}$. 
Next, we use channel-wise multiplication between feature maps and channel attentions to refine the output as:
\begin{equation}
\begin{aligned}
\widetilde{Y_i}=[\widetilde{y_i^{1}},\widetilde{y_i^{2}},&\dots,\widetilde{y_i^{C_i}}]=F_{mul}(Y_i, V_i)
\\ &\widetilde{y_i^{j}}=y_i^{j} \otimes v_i^{j}
\end{aligned}
\end{equation}
where $F_{mul}$ is a function of channel-wise multiplication, $\otimes$ refers to the symbol of channel multiplication and $\widetilde{y_i^{j}}$ denotes the final refined output. 

Through the proposed CRA module, we realize channel attentions to refine the output feature maps,
and then we will introduce the embedding of CRA module into specific CNNs. 

\begin{table*}[!t]
	\caption{The detail configuration of ResNet-50 (Left), SE-ResNet-50 (Middle) and CRA-ResNet-50 (Right).The operations and parameters in residual block are listed in square brackets, The number outside the square brackets denotes the number of stacked residual blocks. The square brackets following by \textit{fc} indicates the output dimension of the two fully connected layers in SE module, The angle brackets following by CRA denotes the configuration of $\langle h_{i},w_{i}\rangle$ in CRA module. Params and FLOPs denotes parameters and floating point operations of the corresponding network, respectively.}
	\label{table:arth_tb}
	\centering
	\begin{tabular}{c|c|c|c}
		\hline
		output size & ResNet-50 & SE-ResNet-50 & CRA-ResNet-50 \\
		\hline
		$\mathrm{112 \times 112}$ & \multicolumn{3}{c}{conv, $7\times7$, 64, stride 2}\\
		\hline
		\multirow{2}*{$56 \times 56 $} &\multicolumn{3}{c}{max pool, $3\times3$, stride 2}\\
		\cline{2-4} & 
		$\begin{bmatrix}
		\begin{array}{l}
		\mathrm{conv},\, 1 \times 1,\, 64 \\
		\mathrm{conv},\, 3 \times 3,\, 64 \\
		\mathrm{conv},\, 1 \times 1,\, 256 \\
		\end{array}
		\end{bmatrix}\times 3$ &
		$\begin{bmatrix}
		\begin{array}{l}
		\mathrm{conv},\, 1 \times 1,\, 64 \\
		\mathrm{conv},\, 3 \times 3,\, 64 \\
		\mathrm{conv},\, 1 \times 1,\, 256 \\
		\mathrm{\textit{fc},\, [16,\, 256]} \\
		\end{array}
		\end{bmatrix}\times 3$ &
		$\begin{bmatrix}
		\begin{array}{l}
		\mathrm{conv},\, 1 \times 1,\, 64 \\
		\mathrm{conv},\, 3 \times 3,\, 64 \\
		\mathrm{conv},\, 1 \times 1,\, 256 \\
		\mathrm{CRA},\, \langle7 \times 7\rangle,\, 256\\
		\end{array}
		\end{bmatrix}\times 3$ \\
		\hline
		${\rm 28 \times 28} $ &
		$\begin{bmatrix}
		\begin{array}{l}
		\mathrm{conv},\, 1 \times 1,\, 128 \\
		\mathrm{conv},\, 3 \times 3,\, 128 \\
		\mathrm{conv},\, 1 \times 1,\, 512 \\
		\end{array}
		\end{bmatrix}\times 4$ &
		$\begin{bmatrix}
		\begin{array}{l}
		\mathrm{conv},\, 1 \times 1,\, 128 \\
		\mathrm{conv},\, 3 \times 3,\, 128 \\
		\mathrm{conv},\, 1 \times 1,\, 512 \\
		\mathrm{\textit{fc},\, [32,\, 512]} \\
		\end{array}
		\end{bmatrix}\times 4$ &
		$\begin{bmatrix}
		\begin{array}{l}
		\mathrm{conv},\, 1 \times 1,\, 128 \\
		\mathrm{conv},\, 3 \times 3,\, 128 \\
		\mathrm{conv},\, 1 \times 1,\, 512 \\
		\mathrm{CRA},\, \langle7 \times 7\rangle,\, 512\\
		\end{array}
		\end{bmatrix}\times 4$ \\
		\hline
		$\mathrm{14 \times 14} $ &
		$\begin{bmatrix}
		\begin{array}{l}
		\mathrm{conv},\, 1 \times 1,\, 256 \\
		\mathrm{conv},\, 3 \times 3,\, 256 \\
		\mathrm{conv},\, 1 \times 1,\, 1024 \\
		\end{array}
		\end{bmatrix}\times 6$ &
		$\begin{bmatrix}
		\begin{array}{l}
		\mathrm{conv},\, 1 \times 1,\, 256 \\
		\mathrm{conv},\, 3 \times 3,\, 256 \\
		\mathrm{conv},\, 1 \times 1,\, 1024 \\
		\mathrm{\textit{fc},\, [64,\, 1024]} \\
		\end{array}
		\end{bmatrix}\times 6$ &
		$\begin{bmatrix}
		\begin{array}{l}
		\mathrm{conv},\, 1 \times 1,\, 256 \\
		\mathrm{conv},\, 3 \times 3,\, 256 \\
		\mathrm{conv},\, 1 \times 1,\, 1024 \\
		\mathrm{CRA},\, \langle7 \times 7\rangle,\, 1024\\
		\end{array}
		\end{bmatrix}\times 6$ \\
		\hline
		${\rm 7 \times 7} $ &
		$\begin{bmatrix}
		\begin{array}{l}
		\mathrm{conv},\, 1 \times 1,\, 512 \\
		\mathrm{conv},\, 3 \times 3,\, 512 \\
		\mathrm{conv},\, 1 \times 1,\, 2048 \\
		\end{array}
		\end{bmatrix}\times 3$ &
		$\begin{bmatrix}
		\begin{array}{l}
		\mathrm{conv},\, 1 \times 1,\, 512 \\
		\mathrm{conv},\, 3 \times 3,\, 512 \\
		\mathrm{conv},\, 1 \times 1,\, 2048 \\
		\mathrm{\textit{fc},\, [128,\, 2048]} \\
		\end{array}
		\end{bmatrix}\times 3$ &
		$\begin{bmatrix}
		\begin{array}{l}
		\mathrm{conv},\, 1 \times 1,\, 512 \\
		\mathrm{conv},\, 3 \times 3,\, 512 \\
		\mathrm{conv},\, 1 \times 1,\, 2048 \\
		\mathrm{CRA},\, \langle7 \times 7\rangle,\, 2048\\
		\end{array}
		\end{bmatrix}\times 3$ \\
		\hline
		$1 \times 1$ & \multicolumn{3}{c}{global average pool, 1000-d \textit{fc}, softmax}\\
		\hline
		$\mathrm{Params} $ & $\mathrm{25.56M}$& $\mathrm{28.09M}$& $\mathrm{26.31M}$\\
		\hline
		$\mathrm{FLOPs}$ & $\mathrm{4.11G}$& $\mathrm{4.12G}$& $\mathrm{4.11G}$\\
		\hline
	\end{tabular}
\end{table*}

\subsection{Network Architectures and Computational Cost}
\label{subsec:architectures}
For CNNs without skip connections, such as VGG \cite{Simonyan2014}, we embed CRA module into convolutional layer directly.
For CNNs with skip connections, such as ResNet \cite{He2016}, we embed CRA module into the last layer of residual block. 
Moreover, the variants of ResNet like ResNeXt \cite{Xie2017} and DenseNet \cite{Huang2017a} can also construct a new architecture in a similar way. 

To show the comparable detail of network configuration, here, we list the architecture configuration of ResNet-50, SE-ResNet-50 and CRA-ResNet-50 in Table.\ref{table:arth_tb}, in which ResNet-50 is taken as the baseline.
It can be seen from the table that CRA module only brings few parameters, and its computational cost is negligible. 
More precisely, in the CRA module, the number of additional floating point operations (FLOPs) is $2C_i(3H_iW_i + h_iw_i)$ and the number of additional parameters is $C_i(h_iw_i+1)$ in the $i$-th layer.

We prefix the original network name with "CRA-" to indicate the network with CRA module and validate their performance in following experiments.

\section{Experiments}
\label{exp}
In this section, we first evaluate CRA module embedded into various baseline networks on the datasets of ImageNet and CIFAR to verify the performance of CRA module in image classification tasks, 
and use Grad-CAM visualization results to show the ability of CRA module on target object regions.
Next, we validate the performance of the proposed method on MS COCO dataset to verify the performance of CRA module in the task of object detection.
Finally, we analyze channel attentions extracted by the proposed CRA module to gain insight into the details of CRA module.

\subsection{Image Classification on ImageNet}
\label{exp_imagenet}
\begin{figure*}[htb]
	\centering
	\includegraphics[width=1.0\linewidth]{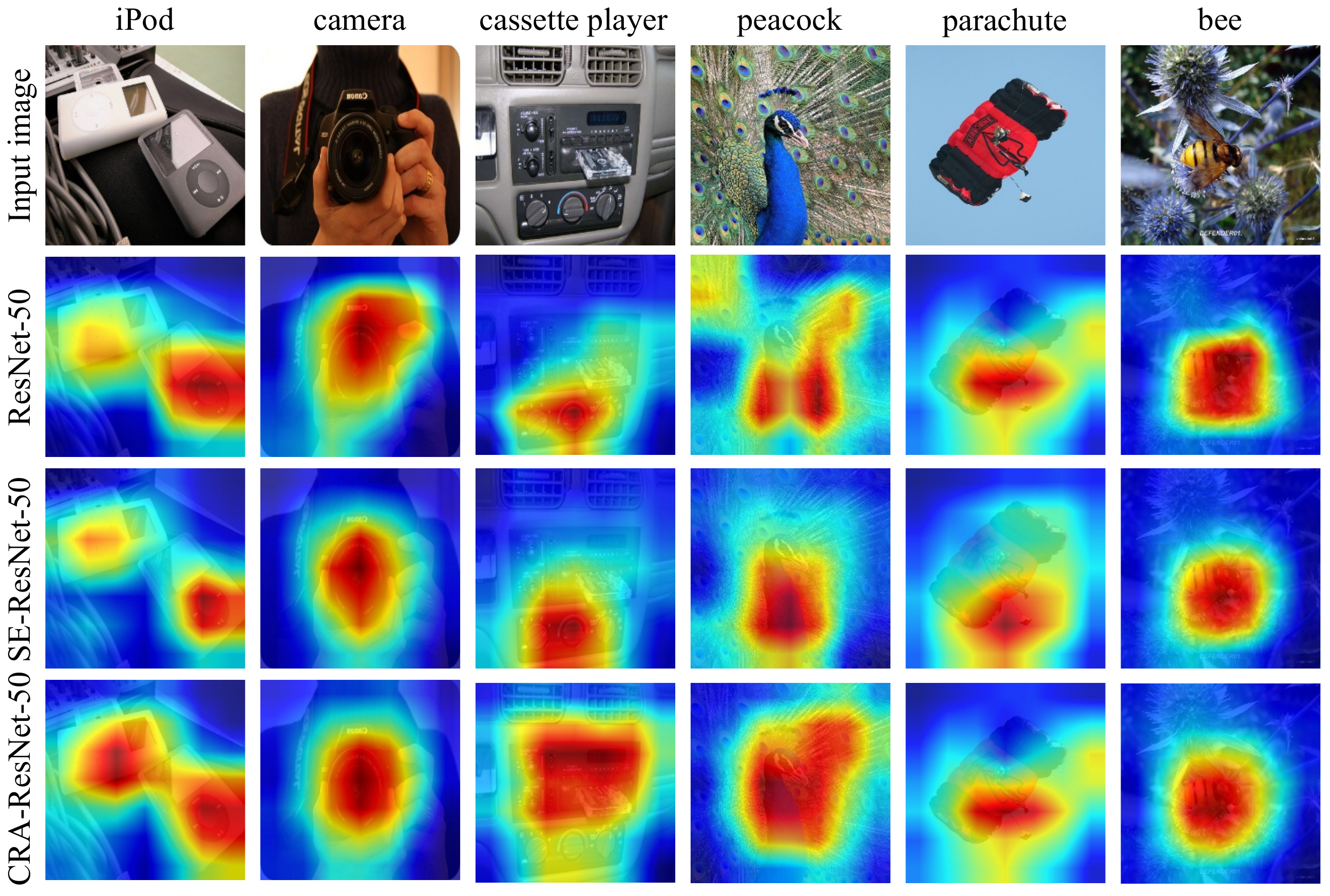}
	\caption{Grad-CAM \cite{Selvaraju2020} visualization results of ResNet-50, SE-ResNet-50 and our CRA-ResNet-50.}
	\label{fig:cam}
\end{figure*}

The ImageNet 2012 classification dataset \cite{JiaDeng2009} contains more than 1.28 million images for training and 50k images for validation from 1000 classes. 
Some sample images of ImageNet 2012 dataset can be seen in the first row in Fig~\ref{fig:cam}.
For the experimental results for all network networks, we report both top-1 and top-5 error rates with a single center crop on the validation set \cite{He2016, Hu2018}.

We follow the practice in \cite{Krizhevsky2012, Simonyan2014, He2016}, initialize the weights as in \cite{He2015} and train 100 epochs totally.
The initial learning rate is 0.1 and divided by 10 every 30 epochs with a batch size of 256. 
The weight decay is 1e-4.
All networks are trained using SGD optimizer with 0.9 momentum \cite{He2016, Hu2018}. 
In order to ensure a fair comparison, we reimplement all the networks with the same settings, including various hyperparameter settings, data augmentation settings, etc.

\begin{table}[!htb]
	\centering
	\caption{Image classification results of CNNs with different arthitectures on ImageNet 2012, single center crop validation errors are reported. }
	\label{tab:exp_imagenet}
	\begin{tabular}{rccccc}
		\toprule
		\multicolumn{2}{c}{\textbf{Architecture}}&\textbf{Top-1 (\%)}&\textbf{Top-5 (\%)}&\textbf{FLOPs}&\textbf{Params}\\
		\hline
		\multicolumn{2}{r}{ResNet-50 \cite{He2016}}&24.20&7.15&4.11G&25.56M\\
		\multicolumn{2}{r}{SE-ResNet-50 \cite{Hu2018}}&23.02&6.65&4.12G&28.09M\\
		\multicolumn{2}{r}{CRA-ResNet-50}&\textbf{22.77}&\textbf{6.47}&4.11G&26.31M\\
		\hline
		\multicolumn{2}{r}{ResNet-101 \cite{He2016}}&23.12&6.67&7.84G&44.55M\\
		\multicolumn{2}{r}{SE-ResNet-101 \cite{Hu2018}}&22.28&6.02&7.85G&49.33M\\
		\multicolumn{2}{r}{CRA-ResNet-101}&\textbf{21.60}&\textbf{5.93}&7.84G&46.17M\\
		\hline
		\multicolumn{2}{r}{ResNeXt-50 \cite{Xie2017}}&22.10&6.14&4.26G&25.03M\\
		\multicolumn{2}{r}{SE-ResNeXt-50 \cite{Hu2018}}&\textbf{21.95}&6.02&4.27G&27.56M\\
		\multicolumn{2}{r}{CRA-ResNeXt-50}&21.99&\textbf{5.87}&4.26G&25.78M\\
		\hline
		\multicolumn{2}{r}{ResNeXt-101 \cite{Xie2017}}&21.27&5.79&8.01G&44.18M\\
		\multicolumn{2}{r}{SE-ResNeXt-101 \cite{Hu2018}}&20.93&5.66&8.03G&48.96M\\
		\multicolumn{2}{r}{CRA-ResNeXt-101}&\textbf{20.71}&\textbf{5.47}&8.02G&45.80M\\
		\toprule
	\end{tabular}
\end{table}

\textbf{Image classification results of different baselines:} 
We carry out the experiments on four baseline networks (i.e., ResNet-50, ResNet-101 \cite{He2016}, ResNeXt-50 and ResNeXt-101 \cite{Xie2017}). 
To compare the performance of CRA module, SE module \cite{Hu2018} is also select to embedded into the above four networks.
The experimental results are shown in Table.\ref{tab:exp_imagenet}. 
It can be seen that the CRA-embedded networks perform much better than both the original networks and SE-embedded networks.
The top-1 errors on both CRA-ResNet-50 (22.77\%) and CRA-ResNet-101 (21.60\%) drop by 1.43\% and 1.52\% respectively compared to that of the original networks.
Even better, our CRA-ResNet-50 has fewer parameters and computational cost than that of ResNet-101 while outperforms it.
Similar to experiments on ResNet, CRA module also brings improvements to ResNeXt obviously.
For the comparison of both accuracy and parameter, CRA module is also superior to SE module. 
The top-1 error of CRA-ResNet-50 is 22.77\% with 26.31M parameters while the top-1 error of SE-ResNet-50 is 23.02\% with 28.09M.
CRA module reduces the parameters by 1.78M but improves the accuracy by 0.25\%, which denotes that the proposed CRA module is efficient.

\begin{figure}
	\centering
	\includegraphics[width=1.0\linewidth]{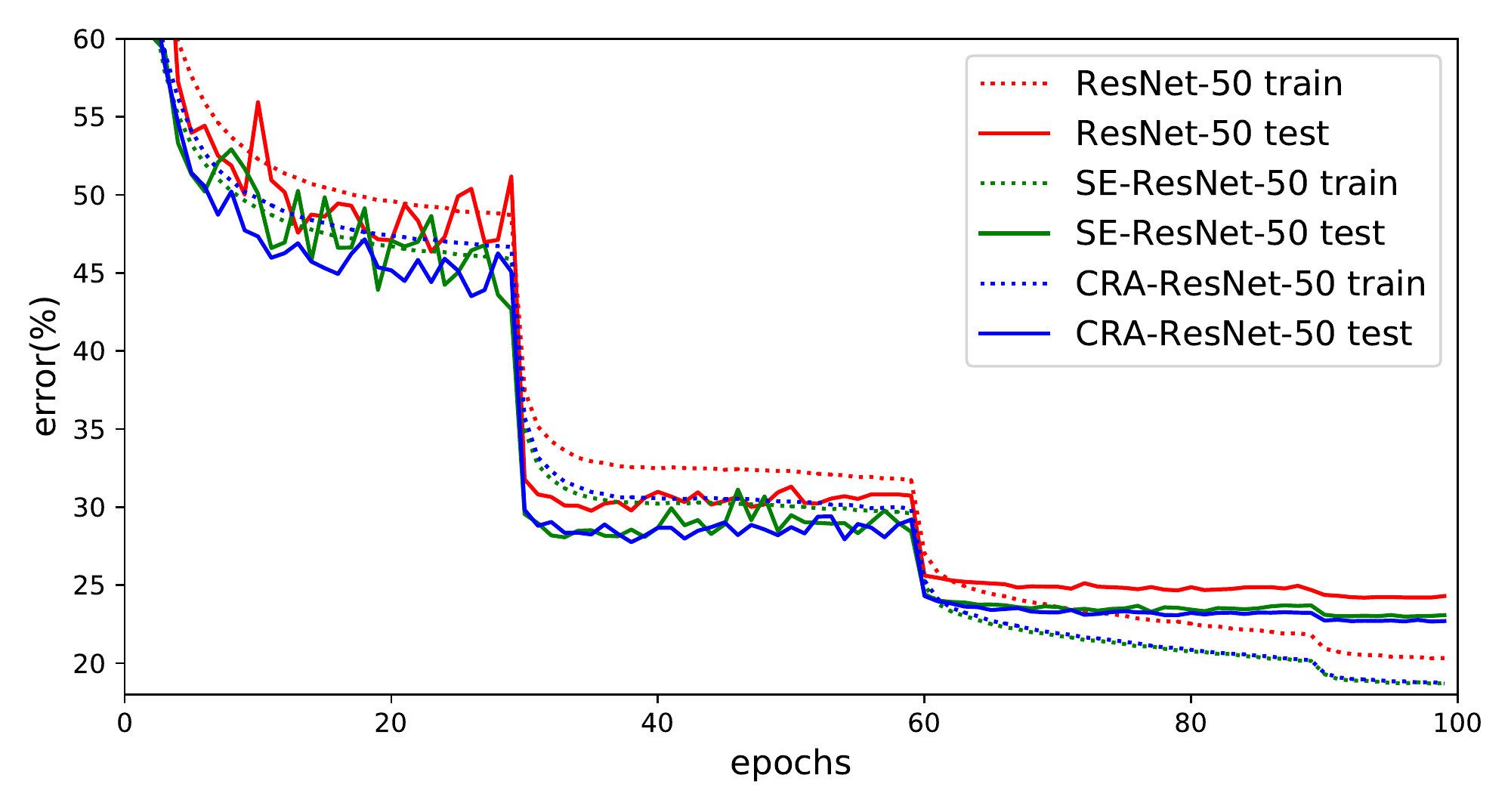}
	\caption{Traning and test error curves of ResNet-50, SE-ResNet-50 and CRA-ResNet-50 on ImageNet 2012.}
	\label{fig:curve1}
\end{figure}

Fig.~$\ref{fig:curve1}$ depicts the training and test error curves of ResNet-50, SE-ResNet-50 and CRA-ResNet-50 during the ImageNet training.
We can observe that the errors of CRA-embedded network are much lower than that of the baselines, and CRA-embedded network is more stable than SE-embedded network during the training process, which indicates that CRA module has stronger fitting ability and representation power.

\begin{table}[!htb]
	\centering
	\caption{Image classification results of ResNet-50 with different attention mechanisms on ImageNet 2012. }
	\label{tab:exp_imagenet2}
	\begin{tabular}{rcccc}
		\toprule
		\multicolumn{1}{c}{{\textbf{Architecture}}}&\textbf{Top-1 (\%)}&\textbf{Top-5 (\%)}&\textbf{FLOPs}&\textbf{Params}\\
		\hline
		ResNet-50 \cite{He2016}&24.20&7.15&4.11G&25.56M\\
		\hline
		BAM \cite{Park2019}&23.21&6.84&4.11G&25.93M\\
		SE \cite{Hu2018}&23.02&6.65&4.12G&28.09M\\
		CBAM \cite{Woo2018}&23.01&6.59&4.13G&28.07M\\
		GALA \cite{Linsley}&22.94&6.52&4.13G&29.42M\\
		CRA&\textbf{22.77}&\textbf{6.47}&4.11G&26.31M\\
		\hline
		\toprule
	\end{tabular}
\end{table}

\textbf{Results of different attention modules:} To compare with other existing attention mechanisms, we benchmark CRA module against BAM \cite{Park2019}, CBAM \cite{Woo2018}, SE \cite{Hu2018}, and GALA \cite{Linsley} on ResNet-50 arthitecture. 
We use the same training strategy as the above for experiments and the results are shown in Table.\ref{tab:exp_imagenet2}.
Among all attention modules, CRA module has the most significant improvement on ResNet-50, and the additional parameters and computational costs introduced by CRA module are moderate.
These results suggest that CRA module provides a competitive trade-off between accuracy and computational cost compared to the previously proposed attention modules.

\textbf{Visualization results of Grad-CAM:} In order to further show the superiority of CRA module, 
we apply the Grad-CAM \cite{Selvaraju2020} to ResNet-50, SE-ResNet-50 and CRA-ResNet-50 for comparison.
Grad-CAM uses the gradients of target concept in convolutional layers to highlight the important areas in the input image for predicting the concept. 
We randomly select some images from ImageNet 2012 validation set and Fig.~$\ref{fig:cam}$ shows the visualization results.
As shown in the figure, the Grad-CAM masks of CRA-ResNet-50 tend to concentrate on more relevant areas in target objects, while the other two networks work obviously not as well as CRA-embedded network. 
Thus, it reveals that CRA module improves the ability of networks to recognize objects.

\begin{table}[tp]
	\centering
	\caption{Image classification results of CRA-ResNet-50 with different $\langle h_i, w_i\rangle$ on ImageNet 2012. }
	\label{tab:exp_diff_hw}
	\begin{tabular}{cccc}
		\toprule
		$\bm{\langle h_i,w_i \rangle}$&\textbf{Top-1 Error (\%)}&\textbf{Top-5 Error (\%)}&\textbf{Params}\\
		\hline
		$\langle 7,7 \rangle$&\textbf{22.77}&\textbf{6.47}&26.31M\\
		$\langle 5,5 \rangle$&22.98&6.58&25.95M\\
		$\langle 3,3\rangle$&23.27&6.53&25.71M\\
		$\langle 1,1\rangle$&23.56&6.88&25.59M\\
		w/o&24.20&7.15&25.56M\\
		\toprule
	\end{tabular}
\end{table}

\textbf{Ablation Study:} According to the previous analysis, $\langle h_i, w_i\rangle$ controls parameters and computational cost of CRA module. 
In order to investigate the effect of different $\langle h_i, w_i\rangle$ on CRA module, we compare the performance of CRA-ResNet-50 with different $\langle h_i, w_i\rangle$ on ImageNet 2012.
CRA-ResNet-50 has $4$ stages as show in Table \ref{table:arth_tb}, when the input image is $224\times224$ pixels, the spatial size of feature maps in the last stage is $7\times7$. Therefore, in our experiments, the maximum value of $\langle h_i, w_i\rangle$ we assigned is $\langle 7, 7\rangle$.
Table.\ref{tab:exp_diff_hw} shows the comparison results under different $\langle h_i, w_i\rangle$.
Note that when $\langle h_i, w_i\rangle$ is set to $\langle 1, 1\rangle$, it is equivalent to using global average pooling for feature maps, that is, spatial information in feature maps is not used.
Table.\ref{tab:exp_diff_hw} shows that spatial information of feature maps has an important influence on the generation of channel attentions,
and indicates that the larger the value of $\langle h_i, w_i\rangle$, the better the performance of CRA-embedded network, because a larger $\langle h_i, w_i\rangle$ can retain more spatial information.

\begin{figure}[!htbp]
	\centering
	\includegraphics[width=1.0\linewidth]{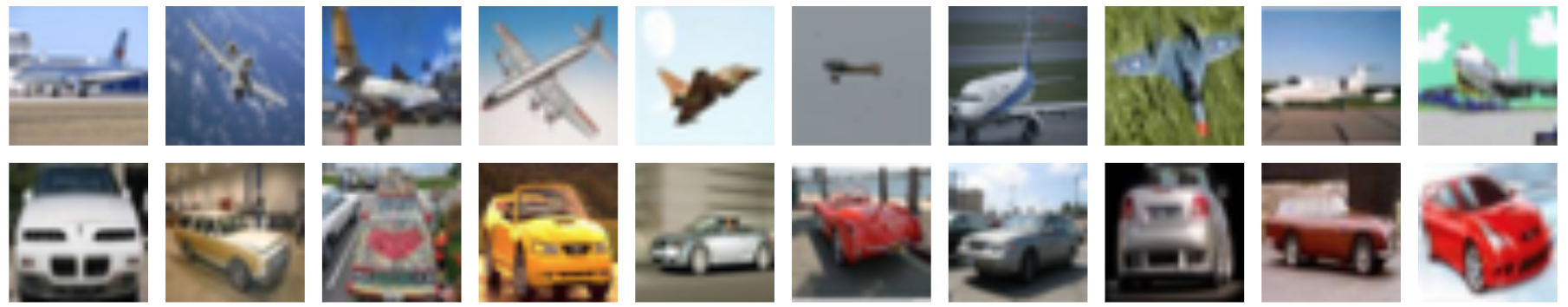}
	\caption{Sample images from CIFAR-10 database.}
	\label{fig:cifar}
\end{figure}

\subsection{Image Classification on CIFAR-10 and CIFAR-100}
CIFAR-10 and CIFAR-100 datasets \cite{Krizhevsky2009} consist 50k images in the training set and 10k images in the test set, and the size of the image is $32 \times 32$ pixels. The images in CIFAR-10 are divided into 10 classes while the ones in CIFAR-100 are divided into 100 classes. Some sample images of CIFAR-10 dataset can be seen in Fig~\ref{fig:cifar}. We report the final top-1 error rates on the test set.

On the two CIFAR datasets, we still conduct experiments with the original networks, SE-embedded networks and CRA-embedded networks.
We use batch size 64 for 300 epochs and follow \cite{He2016} as data augmentation. The initial learning rate is set to 0.1 and divided by 10 at the epoch of 150 and 225. 
According to the different architectures of networks, we set $\langle h_{i},w_{i}\rangle$ to $\langle 8,8\rangle$ for ResNet and ResNeXt, and set it to $\langle 4,4\rangle$ for ShuffleNet \cite{Zhang2018}, DenseNet and VGG in this experiment.

The results are summarized in Table \ref{tab:exp_cifar}. It shows that all CRA-embedded networks perform much better than the other two comparison networks. These results suggest that CRA module brings significant improvements and achieves perfect performance. 

\begin{figure}[H]
	\centering
	\includegraphics[width=1.0\linewidth]{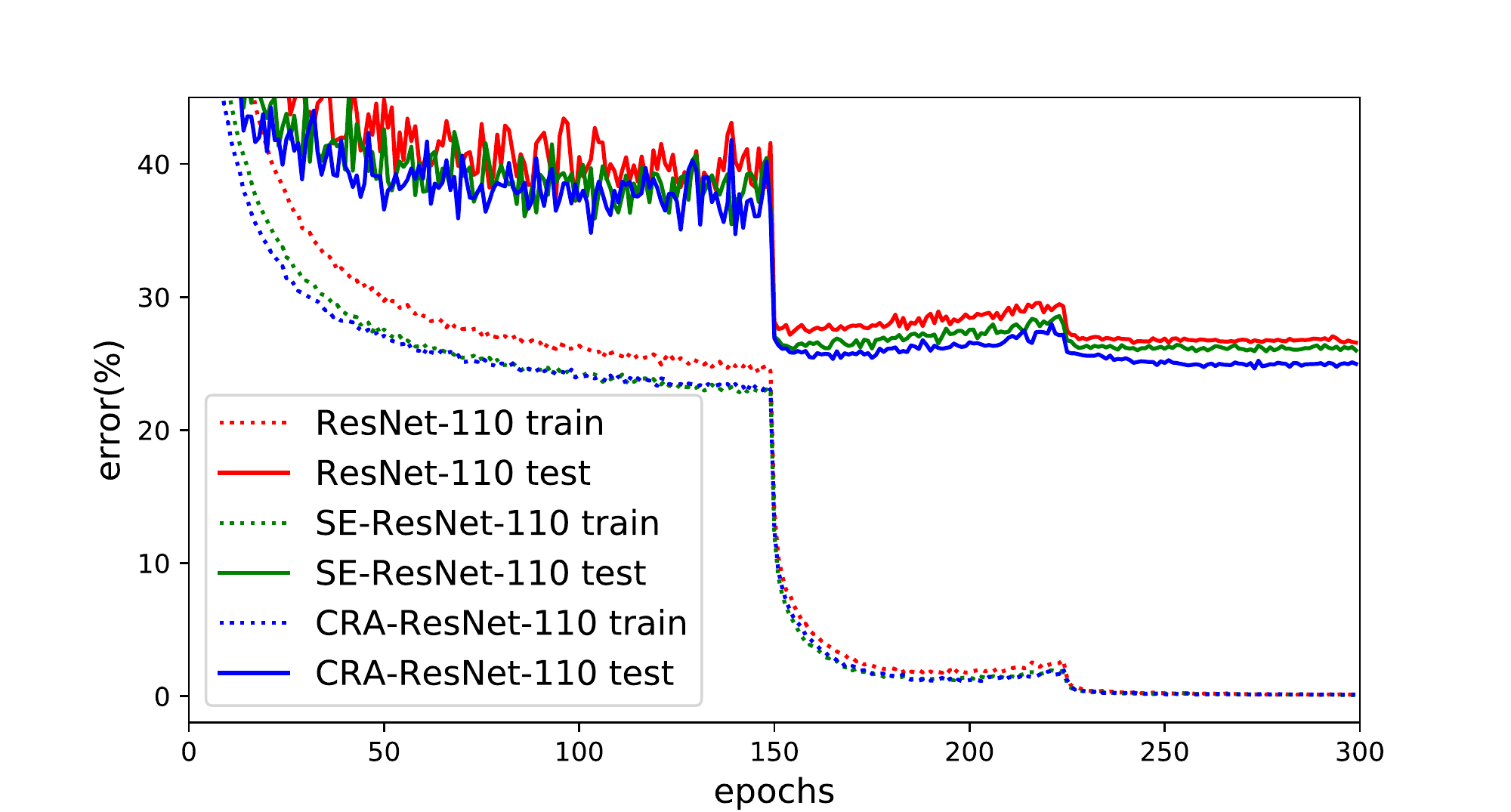}
	\caption{Traning errors and test errors of ResNet-110, SE-ResNet-110 and CRA-ResNet-110 on CIFAR-100.}
	\label{fig:cirve2}
\end{figure}
\begin{table*}[htbp]
	\centering
	\caption{Image classification results of CNNs with different arthitectures on CIFAR-10 and CIFAR-100. Top-1 validation errors are reported.}
	\label{tab:exp_cifar}
	\begin{tabular}{cccrr|crr|crr}
		\toprule
		\multicolumn{2}{c}{\multirow{2}{*}{\diagbox{\textbf{Architecture}}{\textbf{CIFAR-10} }}}&
		\multicolumn{3}{c}{\textbf{Original network}}&	
		\multicolumn{3}{c}{\textbf{SE-embedded network }}&	
		\multicolumn{3}{c}{\textbf{CRA-embedded network}}\\
		\cline{3-11}&&\textit{Error (\%)} & \textit{Params} & \textit{FLOPs} & \textit{Error (\%)} & \textit{Params}& \textit{FLOPs} & \textit{Error (\%)} & \textit{Params}& \textit{FLOPs} \\
		\hline
		\multicolumn{2}{r}{ResNet-56 \cite{He2016}}&6.37&853.02K&126.56M&6.06&860.14K&126.56M&\textbf{5.62}&918.54K&126.62M\\
		\multicolumn{2}{r}{ResNet-110 \cite{He2016}}&5.61&1.73M&254.99M&5.42&1.74M&255.01M&\textbf{5.24}&1.86M&255.12M\\
		\multicolumn{2}{r}{ResNeXt-29 \cite{Xie2017}}&3.74&18.17M&3.03G&3.89&18.69M&3.03G&\textbf{3.73}&18.52M&3.03G\\
		\toprule
		\toprule
		\multicolumn{2}{c}{\multirow{2}{*}{\diagbox{\textbf{Architecture}}{\textbf{CIFAR-100} }}}&
		\multicolumn{3}{c}{\textbf{Original network}}&	
		\multicolumn{3}{c}{\textbf{SE-embedded network }}&	
		\multicolumn{3}{c}{\textbf{CRA-embedded network}}\\
		\cline{3-11}&&\textit{Error (\%)} & \textit{Params} & \textit{FLOPs} & \textit{Error (\%)} & \textit{Params}& \textit{FLOPs} & \textit{Error (\%)} & \textit{Params}& \textit{FLOPs} \\
		\hline
		\multicolumn{2}{r}{ResNet-56 \cite{He2016}}&27.25&858.87K&126.57M&26.74&865.99K&126.58M&\textbf{26.36}&924.39K&126.63M\\
		\multicolumn{2}{r}{ResNet-110 \cite{He2016}}&26.54&1.73M&255.00M&25.91&1.75M&255.02M&\textbf{24.63}&1.86M&255.13M\\
		\multicolumn{2}{r}{ResNeXt-29 \cite{Xie2017}}&18.61&18.26M&3.03G&17.23&18.78M&3.03G&\textbf{17.04}&18.61M&3.03G\\
		\multicolumn{2}{r}{ShuffleNet-1x \cite{Zhang2018}}&28.75&1.01M&44.55M&28.52&1.63M&45.17M&\textbf{27.59}&1.38M&44.92M\\
		\multicolumn{2}{r}{VGG-16 \cite{Simonyan2014}}&26.00&15.30M&315.18M&25.70&15.53M&315.41M&\textbf{25.60}&15.37M&315.25M\\
		\multicolumn{2}{r}{DenseNet-BC\cite{Huang2017a}}&22.66&800.03K&296.59M&22.30&804.21K&296.59M&\textbf{21.64}&950.18K&296.74M\\
		\toprule
	\end{tabular}
\end{table*}

\begin{figure*}[!htb]
	\centering
	\includegraphics[width=.98\linewidth]{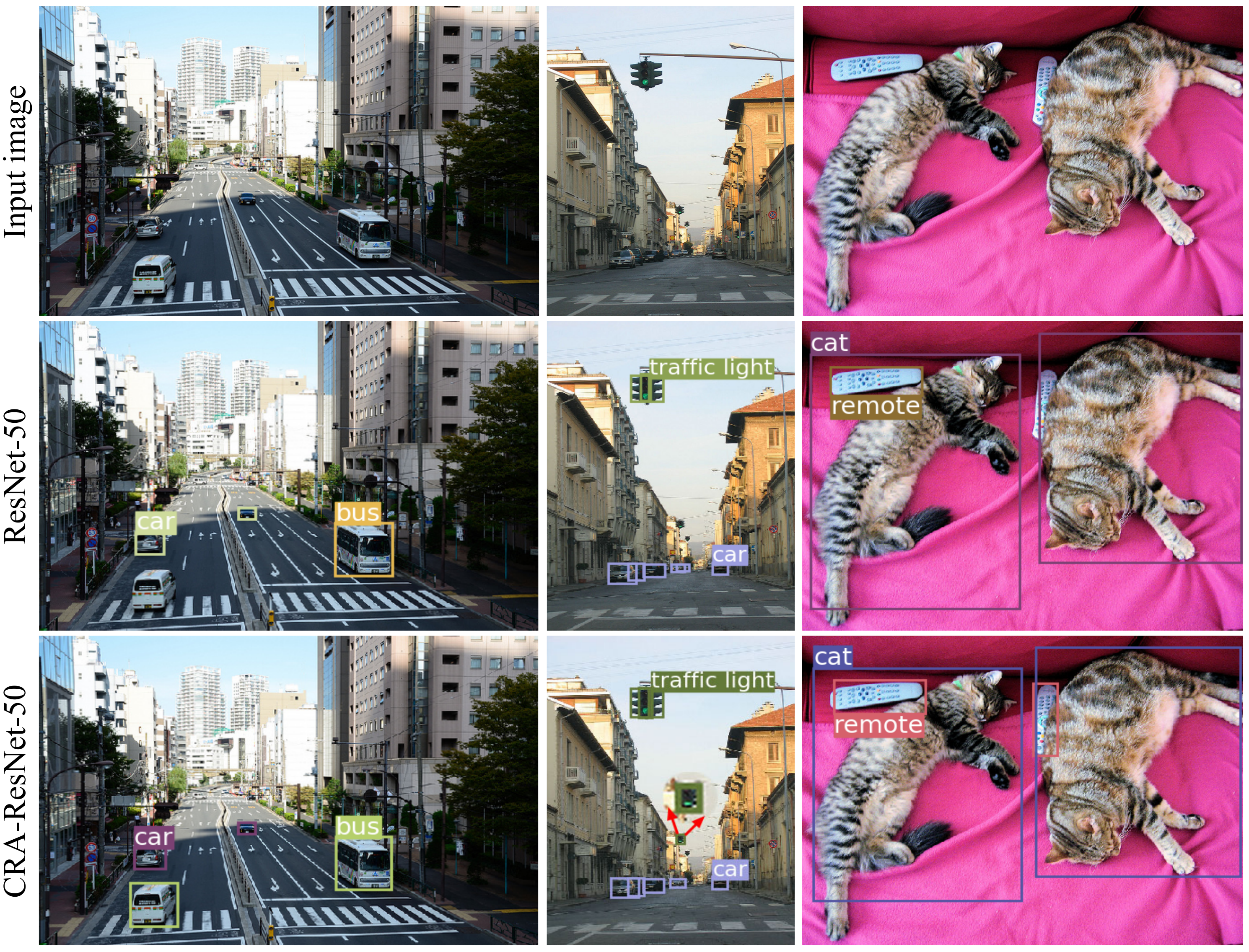}
	\caption{The comparison visualization results of RetinaNet with the original ResNet-50 backbone and our CRA-ResNet-50 backbone on MS COCO dataset. Input images are picked from MS COCO 2017 validation set randomly. 
		The images in the first, second and third row are the original input images, the results of RetinaNet with ResNet-50 backbone, and the results of RetinaNet with CRA-ResNet-50 backbone, respectively.}
	\label{fig:cocovis}
\end{figure*}

\begin{figure*}[!htb]
	\centering
	\includegraphics[width=1.0\linewidth]{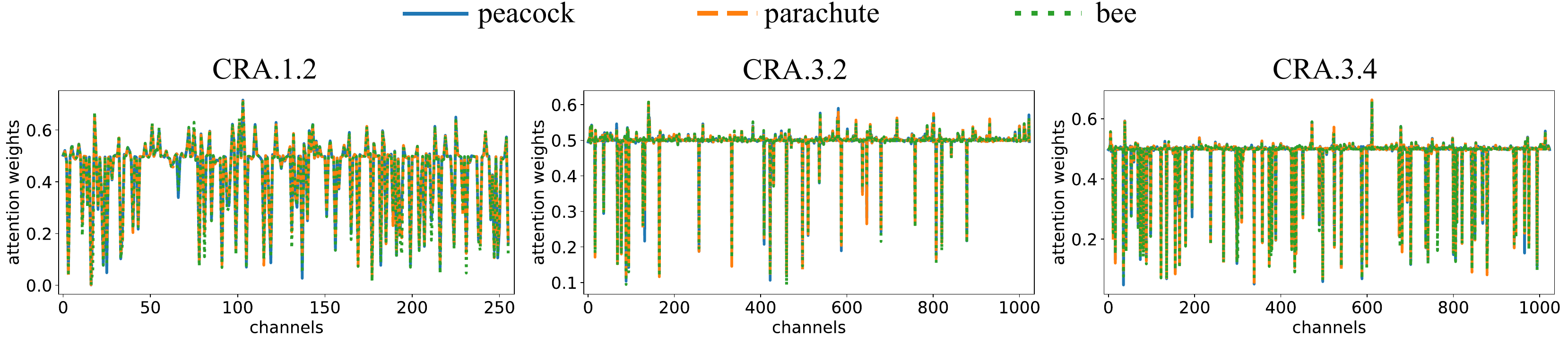}
	\caption{channel attentions extracted by CRA modules in different layers of CRA-ResNet-50 on the dataset of ImageNet 2012. Different style curves represent different input images. CRA modules in different layers are named as "CRA.\textit{stageID.blockID}".}
	\label{fig:CRA_vis}
\end{figure*}
Fig.\ref{fig:cirve2} depicts the training and test curves of ResNet-101, SE-ResNet-101 and CRA-ResNet-101 on CIFAR-100. It is worth noting that after 150 epochs, training errors of three networks tend to be silimar while for the error rates on the test set, CRA-ResNet-101 is obviously superior to ResNet-101 and SE-ResNet-101. 
These curves illustrate that CRA module enhances the information processing ability of feature maps and improves the representation ability of networks.

\subsection{Object Detection on MS COCO}
In order to verify the effectiveness of CRA module on other task, we conduct object detection experiment on MS COCO 2017 dataset \cite{Lin2014}.
MS COCO 2017 is divided into 80 classes, which has more than 118k images in the training set and 5k images in the validation set.
Some sample images of MS COCO 2017 dataset can be seen in the first row in Fig~\ref{fig:cocovis}.
We report the results in mean Average Precision (mAP) over different IoU thresholds from 0.5 to 0.95 on the validation set, following \cite{Lin2017a, Lin2017}.
We use RetinaNet \cite{Lin2017} as our detection framework, ResNet-50, ResNet-101 and their CRA-embedded networks as its backbone respectively. 
These backbones are pre-trained on ImageNet 2012.
We use SGD optimizer and set batch size to 16. 
The initial learning rate is 0.1 and we totally train 90k iterations.
The learning rate is divided by 10 at 60k and 80k iterations with 500 iterations warmup to the initial learning rate.

Table.~\ref{table:coco} shows the experimental results.
As shown in the table, the performance of RetinaNet with CRA-embedded network backbone significantly exceeds that of the original backbone.
This proves that CRA module brings more powerful representation ability and maintains a strong generalization performance.

We analyze the improvement brought by the backbone of CRA-ResNet-50 to RetinaNet from the visualization results, as shown in Fig.~\ref{fig:cocovis}.
It can be found that for the image in the left column, a bus ignored by RetinaNet with ResNet-50 backbone has been detected by RetinaNet with CRA-ResNet-50 based backbone. And for the image in the middle column, a small traffic light is successfully detected by CRA-ResNet-50 backbone based RetinaNet. These indicate that RetinaNet with CRA-ResNet-50 backbone is more sensitive to small and medium objects. 
In addition, for partially occluded objects, such as the remote control in the image in the right column, RetinaNet composed by CRA-embedded backbone can also detect well.
In summary, the visualization results show the good generalization performance of CRA module on object detection tasks.

\begin{table}[!htbp]
	\caption{Object detection results on MS COCO 2017.}
	\label{table:coco}
	\centering
	\begin{tabular}{ccccc}
		\hline
		Backbone & Detector & mAP$_{[.5:.95]}$ & mAP$_{.5}$ & mAP$_{.75}$\\
		\hline
		\multicolumn{1}{r}{ResNet-50 \cite{He2016}}&RetinaNet&33.7\%&52.4\%&36.1\%\\
		\multicolumn{1}{r}{CRA-ResNet-50} &RetinaNet&\textbf{34.2\%}&\textbf{53.6\%}&\textbf{36.4\%} \\
		\hline
		\multicolumn{1}{r}{ResNet-101 \cite{He2016}}&RetinaNet&35.6\%&54.4\%&38.5\%\\
		\multicolumn{1}{r}{CRA-ResNet-101} &RetinaNet&\textbf{36.2\%}&\textbf{55.7\%}&\textbf{38.9\%} \\
		\hline
	\end{tabular}
\end{table}

\subsection{Analysis of Channel Attentions}
To explore the responses of CRA module to different input images, we randomly select 3 images as the inputs of CRA-ResNet-50 (i.e., \textit{peacock}, \textit{parachute} and \textit{bee}, the last three images shown in Fig.\ref{fig:cam}).
Fig.~\ref{fig:CRA_vis} shows channel attentions extracted by CRA modules from different layers in CRA-ResNet-50. We name CRA modules in different layers as "CRA.\textit{stageID.blockID}".

By comparing the responses of CRA modules in different stages (e.g., CRA.1.2 and CRA.3.2), we can observe that channel attentions in shallow layers (CRA.1.2) are much lower than that in deep layers (CRA.3.2).
This is because, as mentioned in \cite{Yosinski2014} \cite{Luo2017}, the shallow layers in networks learn general features, which contain a lot of redundant information, while the deep layers learn the specific features and have more important information. 
For different blocks in the same stage (e.g., CRA.3.2 and CRA.3.4), channel attentions that extracted by CRA module in different channels are different. 
And for different images, channel attention obtained from the same channel tends to be similar, but there are also slight differences.

Through embedding CRA module, the ability of CNNs to capture important information in the forward propagation process is strengthened, and the interference of useless information is reduced, thus the performance of CNNs is improved.

\section{Conclusion}
\label{sec:conclusion}
This paper presents a novel \textit{Channel Reassessment Attention} (CRA) module to improve the performance of networks.
Our module extracts channel attentions through compression and extraction operations based on spatial information of feature maps, and then refines features adaptively.
For image classification tasks, the experimental results on the datasets of ImageNet 2012, CIFAR-10 and CIFAR-100 indicate that CRA module can effectively improve the performance of many different architectures based on CNNs at the minimal additional computational cost.
For object detection tasks, the experimental results on MS COCO 2017 datasets verify the generalization performance of CRA module.

\section*{Acknowledgment}
This work was supported in part by the National Nature Science Foundation of China (61773166), 2030 National Key AI Program of China (2018AAA0100500) and the Science and Technology Commission of Shanghai Municipality under Grant 14DZ2260800.

\newpage
\bibliographystyle{IEEEbib}
\bibliography{references}

\begin{thebibliography}{10}

\bibitem{Krizhevsky2012}
Alex Krizhevsky, Ilya Sutskever, and Geoffrey~E. Hinton,
\newblock ``{ImageNet classification with deep convolutional neural
  networks},''
\newblock in {\em Neural Information Processing Systems (NIPS)}, 2012, pp.
  1097--1105.

\bibitem{Zeiler2014}
Matthew~D. Zeiler and Rob Fergus,
\newblock ``{Visualizing and understanding convolutional networks},''
\newblock in {\em European Conference on Computer Vision (ECCV)}, 2014, pp.
  818--833.

\bibitem{Sermanet2014}
Pierre Sermanet and David Eigen,
\newblock ``{OverFeat : Integrated Recognition , Localization and Detection
  using Convolutional Networks},''
\newblock in {\em International Conference on Learning Representations (ICLR)},
  2014.

\bibitem{Ren2017}
Shaoqing Ren, Kaiming He, Ross Girshick, and Jian Sun,
\newblock ``{Faster R-CNN: Towards Real-Time Object Detection with Region
  Proposal Networks},''
\newblock {\em Neural Information Processing Systems (NIPS)}, pp. 91--99, 2015.

\bibitem{Shelhamer2017}
Evan Shelhamer, Jonathan Long, and Trevor Darrell,
\newblock ``{Fully Convolutional Networks for Semantic Segmentation},''
\newblock in {\em Computer Vision and Pattern Recognition (CVPR)}, 2015, pp.
  3431--3440.

\bibitem{Szegedy}
Christian Szegedy, Wei Liu, Yangqing Jia, Pierre Sermanet, Scott Reed, Dragomir
  Anguelov, Dumitru Erhan, Vincent Vanhoucke, and Andrew Rabinovich,
\newblock ``{Going deeper with convolutions},''
\newblock in {\em Computer Vision and Pattern Recognition (CVPR)}, 2015, pp.
  1--9.

\bibitem{Simonyan2014}
Karen Simonyan and Andrew Zisserman,
\newblock ``{Very Deep Convolutional Networks for Large-Scale Image
  Recognition},''
\newblock {\em International Conference on Learning Representations (ICLR)},
  2015.

\bibitem{He2016}
Kaiming He, Xiangyu Zhang, Shaoqing Ren, and Jian Sun,
\newblock ``{Deep residual learning for image recognition},''
\newblock {\em Computer Vision and Pattern Recognition (CVPR)}, pp. 770--778,
  2016.

\bibitem{Zoph2017}
Barret Zoph and Quoc~V. Le,
\newblock ``{Neural architecture search with reinforcement learning},''
\newblock in {\em International Conference on Learning Representations (ICLR)},
  2017.

\bibitem{Baker2017}
Bowen Baker, Otkrist Gupta, Nikhil Naik, and Ramesh Raskar,
\newblock ``{Designing neural network architectures using reinforcement
  learning},''
\newblock in {\em International Conference on Learning Representations (ICLR)},
  2017.

\bibitem{Ioffe2015}
Sergey Ioffe and Christian Szegedy,
\newblock ``{Batch normalization: Accelerating deep network training by
  reducing internal covariate shift},''
\newblock in {\em International Conference on Machine Learning (ICML)}, 2015,
  pp. 448--456.

\bibitem{Jaderberg2015}
Max Jaderberg, Karen Simonyan, Andrew Zisserman, and Koray Kavukcuoglu,
\newblock ``{Spatial transformer networks},''
\newblock in {\em Neural Information Processing Systems (NIPS)}, 2015, pp.
  2017--2025.

\bibitem{Hu2018}
Jie Hu, Li~Shen, and Gang Sun,
\newblock ``{Squeeze-and-Excitation Networks},''
\newblock in {\em Computer Vision and Pattern Recognition (CVPR)}, 2018, pp.
  7132--7141.

\bibitem{Chen2017}
Long Chen, Hanwang Zhang, Jun Xiao, Liqiang Nie, Jian Shao, Wei Liu, and
  Tat~Seng Chua,
\newblock ``{SCA-CNN: Spatial and channel-wise attention in convolutional
  networks for image captioning},''
\newblock in {\em Computer Vision and Pattern Recognition (CVPR)}, 2017, pp.
  6298--6306.

\bibitem{Park2019}
Jongchan Park, Sanghyun Woo, Joon~Young Lee, and In~So Kweon,
\newblock ``{BAM: Bottleneck Attention Module},''
\newblock in {\em British Machine Vision Conference (BMVC)}, 2018, pp.
  147--160.

\bibitem{Woo2018}
Sanghyun Woo, Jongchan Park, Joon~Young Lee, and In~So Kweon,
\newblock ``{CBAM: Convolutional Block Attention Module},''
\newblock in {\em European Conference on Computer Vision (ECCV)}, 2018, pp.
  3--19.

\bibitem{Wang2018}
Xiaolong Wang, Ross Girshick, Abhinav Gupta, and Kaiming He,
\newblock ``{Non-local Neural Networks},''
\newblock in {\em Computer Vision and Pattern Recognition (CVPR)}, 2018, pp.
  7794--7803.

\bibitem{Fu2019}
Jun Fu, Jing Liu, Haijie Tian, Yong Li, Yongjun Bao, Zhiwei Fang, and Hanqing
  Lu,
\newblock ``{Dual attention network for scene segmentation},''
\newblock in {\em Computer Vision and Pattern Recognition (CVPR)}, 2019, pp.
  3141--3149.

\bibitem{Huang2019}
Zilong Huang, Xinggang Wang, Lichao Huang, Chang Huang, Yunchao Wei, and Wenyu
  Liu,
\newblock ``{CCNet: Criss-cross attention for semantic segmentation},''
\newblock in {\em International Conference on Computer Vision (ICCV)}, 2019,
  pp. 603--612.

\bibitem{Luo2018}
Jian-Hao Luo and Jianxin Wu,
\newblock ``{AutoPruner: An End-to-End Trainable Filter Pruning Method for
  Efficient Deep Model Inference},''
\newblock {\em Pattern Recognition}, vol. 107, pp. 107461, 2020.

\bibitem{Xie2017}
Saining Xie, Ross Girshick, Piotr Doll{\'{a}}r, Zhuowen Tu, and Kaiming He,
\newblock ``{Aggregated residual transformations for deep neural networks},''
\newblock {\em Computer Vision and Pattern Recognition (CVPR)}, pp. 5987--5995,
  2017.

\bibitem{Huang2017a}
Gao Huang, Zhuang Liu, Laurens {Van Der Maaten}, and Kilian~Q. Weinberger,
\newblock ``{Densely connected convolutional networks},''
\newblock in {\em Computer Vision and Pattern Recognition (CVPR)}, 2017, pp.
  2261--2269.

\bibitem{JiaDeng2009}
{Jia Deng}, {Wei Dong}, R.~Socher, {Li-Jia Li}, {Kai Li}, and {Li Fei-Fei},
\newblock ``{ImageNet: A large-scale hierarchical image database},''
\newblock in {\em Computer Vision and Pattern Recognition (CVPR)}, 2009, pp.
  248--255.

\bibitem{He2015}
Kaiming He, Xiangyu Zhang, Shaoqing Ren, and Jian Sun,
\newblock ``{Delving deep into rectifiers: Surpassing human-level performance
  on imagenet classification},''
\newblock in {\em International Conference on Computer Vision (ICCV)}, 2015,
  pp. 1026--1034.

\bibitem{Linsley}
Drew Linsley, Dan Scheibler, Sven Eberhardt, and Thomas Serre,
\newblock ``{Global-and-local attention networks for visual recognition},''
\newblock {\em CoRR, abs/1805.08819, 2018. 5}.

\bibitem{Selvaraju2020}
Ramprasaath~R Selvaraju, Michael Cogswell, Abhishek Das, Ramakrishna Vedantam,
  Devi Parikh, and Dhruv Batra,
\newblock ``{Grad-CAM: Visual Explanations from Deep Networks via
  Gradient-Based Localization},''
\newblock {\em International Conference on Computer Vision (ICCV)}, pp.
  336--359, 2017.

\bibitem{Krizhevsky2009}
Alex Krizhevsky,
\newblock ``{Learning Multiple Layers of Features from Tiny Images},''
\newblock {\em Technical report, University of Toronto}, 2009.

\bibitem{Zhang2018}
Xiangyu Zhang, Xinyu Zhou, Mengxiao Lin, and Jian Sun,
\newblock ``{ShuffleNet: An Extremely Efficient Convolutional Neural Network
  for Mobile Devices},''
\newblock {\em Computer Vision and Pattern Recognition (CVPR)}, pp. 6848--6856,
  2018.

\bibitem{Lin2014}
Tsung~Yi Lin, Michael Maire, Serge Belongie, James Hays, Pietro Perona, Deva
  Ramanan, Piotr Doll{\'{a}}r, and C~Lawrence Zitnick,
\newblock ``{Microsoft COCO: Common objects in context},''
\newblock in {\em European Conference on Computer Vision (ECCV)}, 2014, pp.
  740--755.

\bibitem{Lin2017a}
Tsung~Yi Lin, Piotr Doll{\'{a}}r, Ross Girshick, Kaiming He, Bharath Hariharan,
  and Serge Belongie,
\newblock ``{Feature pyramid networks for object detection},''
\newblock in {\em Computer Vision and Pattern Recognition (CVPR)}, 2017, pp.
  936--944.

\bibitem{Lin2017}
Tsung~Yi Lin, Priya Goyal, Ross Girshick, Kaiming He, and Piotr Dollar,
\newblock ``{Focal Loss for Dense Object Detection},''
\newblock in {\em International Conference on Computer Vision (ICCV)}, 2017,
  pp. 2999--3007.

\bibitem{Yosinski2014}
Jason Yosinski, Jeff Clune, Yoshua Bengio, and Hod Lipson,
\newblock ``{How transferable are features in deep neural networks?},''
\newblock in {\em Neural Information Processing Systems (NIPS)}, 2014, pp.
  3320--3328.

\bibitem{Luo2017}
Jian~Hao Luo, Jianxin Wu, and Weiyao Lin,
\newblock ``{ThiNet: A Filter Level Pruning Method for Deep Neural Network
  Compression},''
\newblock in {\em International Conference on Computer Vision (ICCV)}, 2017,
  pp. 5068--5076.

\end{thebibliography}

%

\end{document}